\def\year{2019}\relax
\documentclass[letterpaper]{article} 
\usepackage{aaai19}  
\usepackage{times}  
\usepackage{helvet}  
\usepackage{courier}  
\usepackage{url}  
\usepackage{graphicx}  
\usepackage{amsmath}
\usepackage{multirow}
\usepackage{array}
\usepackage{adjustbox}
\usepackage{algorithm, algorithmic}
\usepackage{booktabs}

\begin{document}
%
\title{Multi-labeled Relation Extraction with Attentive Capsule Network}

\author{Xinsong~Zhang$^1$,~Pengshuai~Li$^1$,~Weijia~Jia$^{2,1}$\thanks{Corresponding authors: Weijia Jia, Hai Zhao, \{jia-wj, zhaohai\}@cs.sjtu.edu.cn},~and~Hai~Zhao$^1$$^*$\\
$^1$Department of Computer Science and Engineering, Shanghai Jiao Tong University, Shanghai 200240, China\\
$^2$State Key Lab of IoT for Smart City, University of Macau, Macau 999078, China\\
 {\tt \{xszhang0320, pengshuai.li\}@sjtu.edu.cn~and~\{jia-wj, zhaohai\}@cs.sjtu.edu.cn} \\
}

\maketitle
\begin{abstract}
To disclose overlapped multiple relations from a sentence still keeps challenging. Most current works in terms of neural models inconveniently assuming that each sentence is explicitly mapped to a relation label, cannot handle multiple relations properly as the overlapped features of the relations are either ignored or very difficult to identify. To tackle with the new issue, we propose a novel approach for multi-labeled relation extraction with capsule network which acts considerably better than current convolutional or recurrent net in identifying the highly overlapped relations within an individual sentence. To better cluster the features and precisely extract the relations, we further devise attention-based routing algorithm and sliding-margin loss function, and embed them into our capsule network. The experimental results show that the proposed approach can indeed extract the highly overlapped features and achieve significant performance improvement for relation extraction comparing to the state-of-the-art works.
\end{abstract}

\section{Introduction}
Relation extraction plays a crucial role in many natural language processing (NLP) tasks. It aims to identify relation facts for pairs of entities in a sentence to construct triples like [\emph{Arthur Lee, place\_born, Memphis}]. Relation extraction has received renewed interest in the {\it neural network} era, when neural models are effective to extract semantic meanings of relations. Compared with traditional approaches which focus on manually designed features, neural methods such as Convolutional Neural Network (CNN)~\cite{liu2013convolution,zeng2014relation} and Recurrent Neural Network (RNN)~\cite{zhang2015relation,zhou2016attention} have achieved significant improvement in relation classification. However, previous neural models are unlikely to scale in the scenario where a sentence has multiple relation labels and face the challenges in extracting highly overlapped and discrete relation features due to the following two drawbacks.

\begin{figure*}[htbp]
\centering
\includegraphics[width=6in]{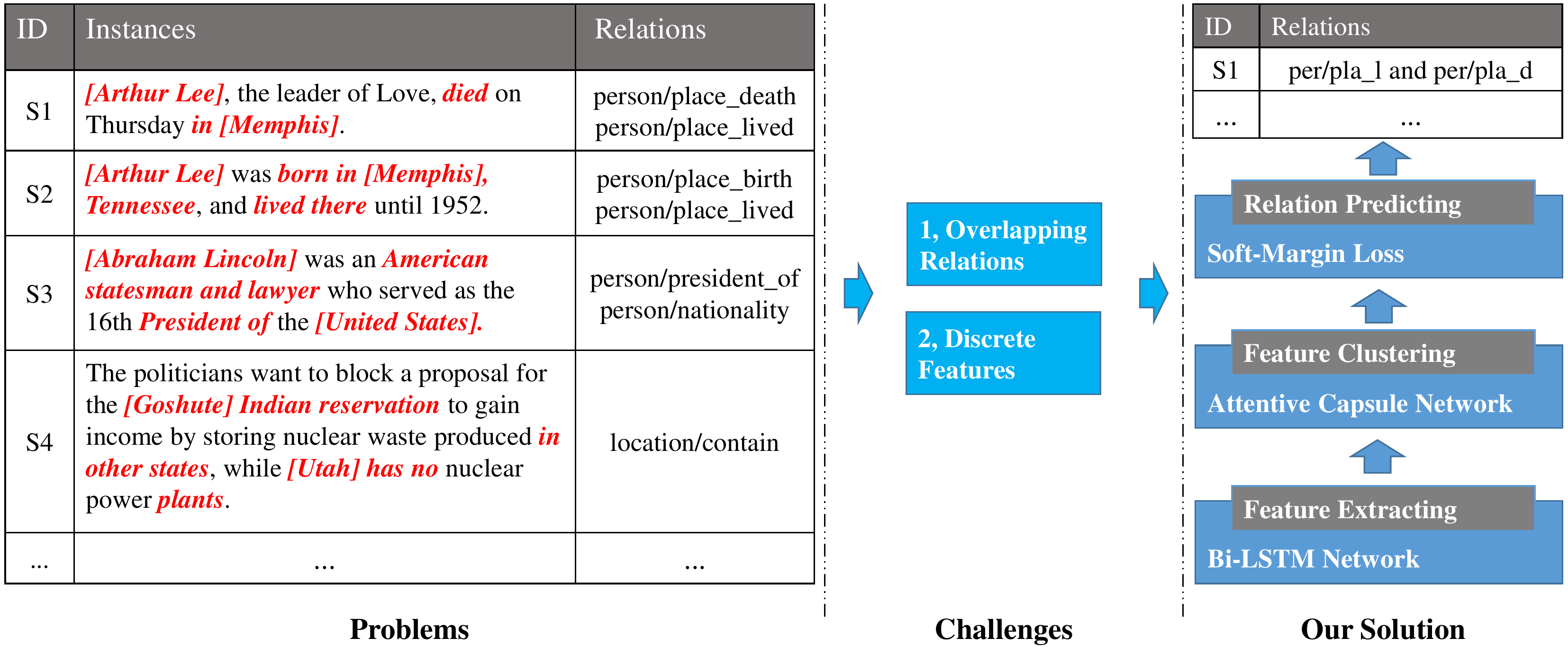}
\caption{Problems, challenges and our solution for multi-labeled relation extraction. Words in brackets are entities and the italic red parts are key words that contain relation features. (The relation label in the right table is in abbreviation.)}
\label{fig:overview}
\end{figure*}

First, one entity pair can express multiple relations in a sentence, which will confuse relation extractor seriously. For example, as in Figure~\ref{fig:overview}, the entity pair [\emph{Arthur Lee, Memphis}] keeps three possible relations which are \emph{place\_birth}, \emph{place\_death} and \emph{place\_lived}. The sentence \emph{S1} and \emph{S2} can both express two relations, and the sentence \emph{S3} represents another two relations. These sentences contain multiple kinds of relation features which are difficult to be identified clearly. The existing neural models tendentiously merge low-level semantic meanings to one high-level relation representation vector with methods such as max-pooling~\cite{zeng2014relation,zhang2016probabilistic} and word-level attention~\cite{zhou2016attention}. However, one high-level relation vector is still insufficient to express multiple relations precisely.

Second, current methods are neglecting of the discretization of relation features. For instance, as shown in Figure~\ref{fig:overview},  all the sentences express their relations with a few significant words (labeled italic in the figure) distributed discretely in the sentences. However, common neural methods handle sentences with fixed structures, which are difficult to gather relation features of different positions. For example, being spatially sensitive, CNNs adopt convolutional feature detectors to extract local patterns from a sliding window of vector sequences and use the max-pooling to select the prominent ones. Besides, the feature distribution of ``no relation (NA, others)'' in a dataset is different from that of definite relations. A sentence can be classified to ``no relation'' only when it does not contain any features of other relations.

In this paper, to extract overlapped and discrete relation features, we propose a novel approach for multi-labeled relation extraction with an attentive capsule network. As shown in Figure~\ref{fig:overview}, the relation extractor of the proposed method is constructed with three major layers that are feature extracting, feature clustering and relation predicting. The first one extracts low-level semantic meanings. The second layer clusters low-level features to high-level relation representations, and the final one predicts relation types for each relation representation. The low-level features are extracted with traditional neural models such as Bidirectional Long Short-Term Memory (Bi-LSTM) and CNN. For the feature clustering layer, we utilize an attentive capsule network inspired by~\citeauthor{sabour2017dynamic}~\shortcite{sabour2017dynamic}. Capsule (vector) is a small group of neurons used to express features. Its overall length indicates the significance of features, and the direction of a capsule suggests the specific property of the feature. The low-level semantic meanings from the first layer are embedded to amounts of low-level capsules, which will be routed and clustered together to represent high-level relation features. For better relation extraction, we further devise an attention-based routing algorithm to precisely find low-level capsules that contain related relation features. Besides, we propose a sliding-margin loss function to address the problem of ``no relation'' in multiple labels scenario. A sentence is classified as ``no relation'' only when the probabilities for all the other specific classes are below a boundary. The boundary is dynamically adjusted in the training process. Experimental results on two widely used benchmarks show that the proposed method can significantly enhance the performance of relation extraction. The contributions of this paper can be summarized as follows,
\begin{itemize}
\item We first apply capsule network to multi-labeled relation extraction by clustering relation features.
\item We propose an attention-based routing algorithm to precisely extract relation features and a sliding-margin loss function to well learn multiple relations.
\item Our experiments on two benchmarks show our method gives new state-of-the-art performance.
\end{itemize}

\section{Related work}
\noindent{\bf Relation Extraction.} Relation extraction is a critical task for the NLP in which supervised methods with human-designed features have been well studied~\cite{bunescu2005subsequence,mintz2009distant,riedel2010modeling,surdeanu2012multi}. Recent years, neural models are widely used to remove the inconvenience of hand-crafted feature design. Both CNN and RNN have been well applied to relation extraction~\cite{socher2012semantic,liu2013convolution,kim2014convolutional,zeng2014relation,santos2015classifying,zhang2015relation}. From the CNN or RNN backbone, relation extraction can be further improved by integrating attention mechanism~\cite{wang2016relation,lin2016neural,zhou2016attention,zhu2017relation}, parser tree~\cite{xu2015semantic,miwa2016end,xu2016improved}, multi-task learning~\cite{liu2016recurrent} or ensemble models~\cite{nguyen2015combining,yang2018ensemble}. However, all the previous neural models simply represent relation features with one vector, resulting in unacceptable precisions for multi-labeled relation extraction.

\noindent{\bf Capsule Network.} Capsule network was proposed to improve the representational limitations of CNN and RNN~\cite{hinton2011transforming}. Capsules with transformation matrices allow networks to learn part-whole relationships automatically. Consequently, a dynamic routing algorithm~\cite{sabour2017dynamic} was proposed to replace the max-pooling in CNN, which achieved impressive performance recognizing highly overlapping digits. Then,~\citeauthor{xi2017capsule}~\shortcite{xi2017capsule} further tested out the application of capsule networks on the CIFAR data with higher dimensionality. \citeauthor{hinton2018matrix}~\shortcite{hinton2018matrix} proposed a new routing method between capsule layers based on the EM algorithm. Recently, capsule network was applied to NLP tasks such as text classification~\cite{zhao2018investigating} and disease classification~\cite{wang2018recurrent}.

Different from the previous methods for multi-labeled relation extraction, we first introduce the capsule network to the task, especially, with two highlighted improvements, attentive routing algorithm and sliding-margin loss.

\section{Method}

\begin{figure*}[htbp]
\centering
\includegraphics[width=6in]{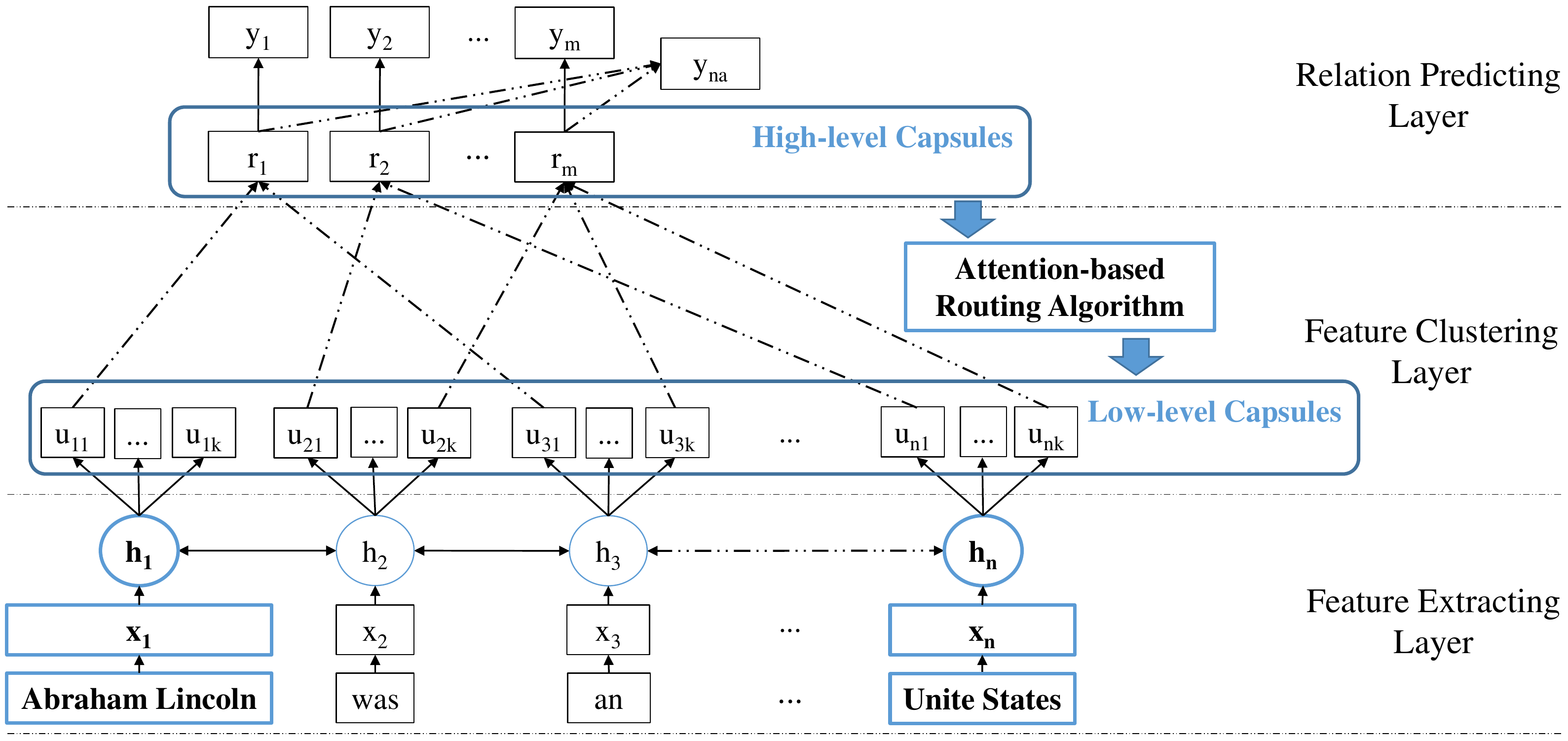}
\caption{The architecture of our proposed relation extractor, illustrating the procedure for handling one sentence and predicting possible relations between [{\it Abraham Lincoln}] and [{\it Unite States}]. $h$ is a set of hidden states of Bi-LSTM, $u$ is a low-level capsule set and $r$ represents high-level capsules. $y$ indicates relation labels, and $y_{na}$ expresses ``no-relation''. The solid lines are determinate associations, and the dotted lines are possible ones.}
\label{fig:architecture}
\end{figure*}

This section describes our approach for multi-labeled relation extraction with an attentive capsule network. As shown in Figure~\ref{fig:architecture}, our relation extractor comprises of three primary layers,
\begin{itemize}
\item {\bf Feature Extraction Layer.} Given a sentence $b^{\ast}$ and two target entities, a Bi-LSTM network is used to extract low-level features of the sentence.
\item {\bf Feature Clustering Layer.} Given vectors of low-level features, we cluster the related features into a high-level relation representation for each relation with an attentive capsule network. 
\item {\bf Relation Predicting Layer.} After computing the high-level representations of a sentence, we apply a sliding-margin loss function to predict possible relations including the label ``no relation''.
\end{itemize}

\subsection{Feature Extracting Layer}
This layer encodes word tokens and extracts low-level semantic information.

\subsubsection{Input Representation}
The input representations of our model include word embeddings and position embeddings.

\noindent{\bf Word Embeddings} are distributed representations of words that map each word to a $p$ dimensional real-valued vector $word$. The vectors are pre-trained in the $skip-gram$ setting of $word2vec$~\cite{mikolov2013efficient}. 

\noindent{\bf Position Embeddings} are defined as the combination of the relative distances from the current word to the entities. For instance, in sentence \emph{Arthur Lee was born in Memphis.}, the relative distances from the word \emph{born} to [\emph{Arthur Lee}] and [\emph{Memphis}] are respectively 2 and -2. We also encode distances to vectors $position \in \mathcal R^q $, where $q$ is the dimension of position embeddings. In our work, the position embeddings are initialized randomly.

The word embeddings and position embeddings are concatenated together as network input vector. We denote all the words in a sentence as an initial vector sequence $b^{\ast} = \{x_1,\cdots,x_i,\cdots,x_n\}$, where $x_i \in  \mathcal R^{p+q}$ and $n$ is the word number.

\subsubsection{Bidirectional LSTM}
Recurrent Neural Network (RNN) is powerful to model sequential data and has achieved great success in the relation classification~\cite{zhang2015relation,zhou2016attention}. RNN for our task is implemented as LSTM with four components~\cite{graves2013generating}, one input gate $i_t$ with corresponding weight matrix $W_i, U_i,V_i$, one forget gate $f_t$ with corresponding weight matrix $W_f, U_f,V_f$, one output gate $o_t$ with corresponding weight matrix $W_o, U_o,V_o$ and one cell $c_t$ with corresponding weight matrix $W_c, U_c,V_c$. All of those components are set to generate the current hidden state $h_t$ with the input token $x_t$ and the previous hidden state $h_{t-1}$. The whole procedure is demonstrated with the following equations,
\begin{equation}
\nonumber
\begin{split}
& i_t = \sigma (W_i[x_t]+U_ih_{t-1}+V_ic_{t-1}+b_i) \\  
& f_t = \sigma (W_f[x_t]+U_fh_{t-1}+V_fc_{t-1}+b_f) \\ 
& c_t = i_t{\tanh}(W_c[x_t]+U_ch_{t-1}+V_cc_{t-1}+b_c)+f_tc_{t-1} \\ 
& o_t = \sigma (W_o[x_t]+U_oh_{t-1}+V_oc_t+b_o) \\ 
& h_t = o_t {\tanh} (c_t),
\end{split}
\end{equation}
where $\sigma$ is the sigmoid function and all of the components have the same sizes as the hidden vector $h$.

For many sequence modeling tasks, it is beneficial to have access to the future context as well as the past. Bidirectional-LSTM network extends the standard LSTM network by introducing a second layer, in which the hidden states flow in an opposite temporal order. The model is, therefore, able to exploit features both from the past and the future. Consequently, we use Bi-LSTM including both forward subnetwork and backward subnetwork to capture global sequence information. The final state of $h_t$ is shown by the equation $h_t = [\overrightarrow {h_t} \oplus \overleftarrow {h_t}]$, where $\overrightarrow {h_t}$ is the forward state, $\overleftarrow {h_t}$ is the backward state and $\oplus$ is the element-wise sum. The dimension of the hidden state is determined by a hyper-parameter $s_h$.

\subsection{Feature Clustering Layer}
This layer clusters features with the help of an attention-based routing algorithm.

\subsubsection{Feature Clustering with Capsule Network}
Capsule network has been proved effective in digital recognition, especially for the highly overlapping digits~\cite{sabour2017dynamic}. In our work, a capsule is a group of neurons whose activity vector represents the instantiation parameters of a specific type of relation features. The length of the activity vector represents the probability that the relation features exist, and the orientation of the vector expresses the specific property of one kind of features. Active capsules make predictions, via transformation matrices, for the instantiation parameters of higher-level capsules. While, high-level capsules are clustered from low-level capsules, which contain local and trivial features. When multiple predictions agree, a higher level capsule becomes active. 

For the task of relation extraction, we scatter all the low-level semantic information extracted by the Bi-LSTM into amounts of low-level capsules represented by $u \in \mathcal R^{d_u}$. The representation of each word token will be expressed by $k$ low-level capsules. Each low-level capsule is applied with a nonlinear squash function $g$ through the entire vector,
\begin{equation}
\nonumber
\begin{split}
& h_t = [u^{'}_{t1};\cdots;u^{'}_{tk}] \\
& u_{tk} = g (u^{'}_{tk})=\frac{{||u^{'}_{tk}||}^2}{1+{||u^{'}_{tk}||}^2} \frac{u^{'}_{tk}}{||u^{'}_{tk}||},\\
\end{split}
\end{equation}
where $[x;y]$ denotes the vertical concatenation of $x$ and $y$. Amounts of low-level capsules can be clustered together to represent high-level relation features. Therefore, high-level capsules $r \in \mathcal R^{d_r}$ are computed with the following equations,
\begin{equation}
\nonumber
	r_{j}= g(\sum_{i} w_{ij}W_{j}u_{i}),\\
\end{equation}
where $w_{ij}$ are coupling coefficients that are determined by an iterative dynamic routing process and $W_{j} \in \mathcal R^{d_{r} \times d_{u}}$ are weight matrices for each high-level capsule. 

\subsubsection{Attention-based Routing Algorithm}
With the capsule network, we can obtain high-level capsules which represent relation features. However, the traditional dynamic routing algorithm in~\cite{sabour2017dynamic} does not focus on the entity tokens, which have been proved important for relation extraction~\cite{wang2016relation,liu2018stp}. Therefore, we propose an attention-based routing algorithm which focuses on the entity tokens when routing for related low-level capsules as in Algorithm~\ref{alg:abra}. The coupling coefficients $w$ between $i$-th capsule and all the capsules in the $r$ sum to 1 and are determined by a ``softmax'' function whose initial logits are $b_{ij}$, the $\log$ prior probabilities that capsule $u_{i}$ should be coupled to capsule $r_{j}$. Besides, we propose attention weights $\alpha$ for all the low-level capsules to maximize the weights of capsules from significant word tokens and minimize that of irrelevant capsules. The weight of the capsule $u_{i}$ is computed by the entity features $h_{e}$ and hidden state $h_{t}^i$ from which the $u_{i}$ comes. The $w$ and $\alpha$ are computed by

\begin{equation}
\nonumber
\begin{split}
	& w_{ij} = \frac{\exp(b_{ij})}{\sum_{j^{\ast}} \exp (b_{ij^{\ast}})}\\
	& \alpha_{i} = \sigma(h_{e}^Th_{t}^i), \\
\end{split}
\end{equation}
where $h_{e}$ is the sum of hidden states of the two entities and $T$ means the transpose operation. The sigmoid function normalizes the attention weights and maximizes the differences between  significant capsules and irrelevant ones. Finally, the high-level capsules $r$ are computed with the Algorithm~\ref{alg:abra}.

\begin{algorithm}
\caption{Attention-based Routing Algorithm}
\label{alg:abra}
\begin{algorithmic}[1]
\REQUIRE low-level capsules $u$, iterative number $z$, entity features $h_{e}$ and hidden states $h_{t}$
\ENSURE high-level capsules $r$
\FOR{all capsules $u_{i}$ and capsules $r_{j}$}
	\STATE initialize the logits of coupling coefficients
	\STATE $b_{ij}$ = 0
\ENDFOR
\FOR {z iterations} 
	\STATE $w_{i}$ =  softmax($b_{i}$), $\forall u_{i} \in u$
	\STATE $\alpha_{i}= \sigma(h_{e}^Th_{t}^i)$, $\forall u_{i} \in u$
	\STATE $r_{j}= g(\sum_{i} w_{ij}\alpha_{i}W_{j}u_{i})$, $\forall r_{j} \in r$
	\STATE $b_{ij} = b_{ij} + W_{j}u_{i}r_{j}$, $\forall u_{i} \in u$ and $\forall r_{j} \in r$
\ENDFOR
\end{algorithmic}
\end{algorithm}

\subsection{Relation Predicting Layer}
In the capsule network, the length of the activity high-level capsules can represent the probability of relations. \citeauthor{sabour2017dynamic}~\shortcite{sabour2017dynamic} applied a fixed margin loss for the classification of digit images. However, they can only set the margin empirically. Therefore, we propose a sliding-margin loss for the task of relation extraction, which learns the baseline of the margin automatically. Besides, to deal with the ``no relation'' (presented as NA in the figure~\ref{fig:architecture}) properly, the probabilities of all the existing relations for sentences labeled as NA should be under the lower bound of the margin. The loss function for the $j$-th relation follows the below equation,
\begin{equation}
\nonumber
\begin{split}
	L_{j} = Y_{j}max(0, (B + \gamma) - ||r_{j}||)^{2} + \\ 
			\lambda (1-Y_{j})max(0, ||r_{j}|| - (B - \gamma))^{2}, \\
\end{split}
\end{equation}
where $Y_{j} = 1$ if the sentence represents relation $r_{j}$, and $Y_{j} = 0$ if not. $\gamma$ is a hyper-parameter defining the width of the margin, and $B$ is a learnable variable indicating the NA threshold of the margin, which is initialized by 0.5. $\lambda$ is the down-weighting of the loss for absent relations, which is the same as that in~\cite{sabour2017dynamic}. The total loss of a sentence is the sum of losses from all the relations. In the testing process, relation labels will be assigned to a sentence when its probabilities of these relations are larger than the threshold $B$. Otherwise, it will be predicted as NA.

\section{Experiments}
We conduct experiments to answer the following three questions. 1) Does our method outperform previous works in relation extraction? 2) Is attentive capsule network useful to distinguish highly overlapping relations? 3) Are the two proposed improvements both effective for relation extraction? 

\subsection{Dataset, Evaluation Metric and Baselines}
\noindent{\bf Dataset}. We conduct experiments on two widely used benchmarks for relation extraction, NYT-10~\cite{riedel2010modeling} and SemEval-2010 Task 8 dataset~\cite{hendrickx2009semeval}. The NYT-10 dataset is generated by aligning Freebase relations with the New York Times (NYT) corpus, in which sentences from the years 2005-2006 are for training while those from 2017 for testing. The dataset consists of amounts of multi-labeled sentences. The SemEval-2010 Task 8 dataset is a small dataset which has been well-labeled for relation extraction. The details of both datasets are shown in Table~\ref{table:dataset}.

\begin{table}[htb]
\centering
\resizebox{\linewidth}{!}{
\begin{tabular}{ccccc}
\toprule
Datasets & Train Sen. & Test Sen. & Multi-labeled Sen. &  Classes \\
\midrule
NYT-10 & 566,190 & 170,866 & 45,693 &  53 \\
Sem. & 8,000 & 2,717 & 0 &  19 \\
\bottomrule
\end{tabular}
}
\caption{Detail information for datasets. {\bf Sen.} is the number of sentences. {\bf Sem.} represents SemEval-2010 Task 8.}
\label{table:dataset}
\end{table}

\noindent{\bf Evaluation Metric}. We evaluate our method with a classical held-out evaluation for NYT-10 and macro-averaged F1 for SemEval-2010 Task 8. The held-out evaluation evaluates our models by comparing the relation facts discovered from the test articles with those in Freebase, which provides an approximation of the precision without the time-consuming human evaluation. Besides, we report both the aggregate Precision-Recall (PR) curves and macro-averaged F1 as quantitative indicators.

\noindent{\bf Baselines}. We select following feature clustering methods as baselines.

{\bf Max-pooling+CNN} clusters the relations extracted by CNN with max-pooling~\cite{zeng2014relation}. 

{\bf Max-pooling+RNN} clusters the relations extracted by RNN with max-pooling~\cite{zhang2015relation}.

{\bf Avg+RNN} aggregates the relation features with linear average of all the hidden states of word tokens.

{\bf Att+RNN} applies a word-level attention to aggregate relation features instead of linear average~\cite{zhou2016attention}. 

{\bf Att-CapNet (CNN-based)} integrates our attentive capsule network with a CNN relation extractor.

{\bf Att-CapNet (RNN-based)} is our method.

\subsection{Experimental Settings}
In our experiments, word embeddings are pre-trained with the $word2vec$ tool. For NYT-10, we pre-train the word embeddings on NYT-10 in the $skip-gram$ setting. In order to compare with the previous works on SemEval-2010 Task 8, we use the same word vectors proposed by~\cite{turian2010word} (50-dimensional) to initialize the embedding layer. Additionally, we also use the 100-dimensional word vectors pre-trained in $Glove$ setting~\cite{pennington2014glove}. Besides, we concatenate the words of an entity when it has multiple words. Position embeddings are initialized randomly and updated in training. We use Adam optimizer~\cite{kingma2014adam} to minimize the objective function. $L2$ regularization and dropout~\cite{srivastava2014dropout} are adopted to avoid overfitting. To train our model efficiently, we iterate by randomly selecting a batch from the training set until convergence. We use a grid search to determine the optional parameters. Table~\ref{table:parameters} lists our hyper-parameter setting\footnote{The parameters of the baselines are following their papers.}.

\begin{table}[htb]
\centering
\normalsize
\begin{tabular}{ccc}
\toprule
Parameters & NYT-10 & Sem. \\ 
\midrule
batch size  & 50 & 50\\
word dimension $p$ & 50  & 50 \\
position dimension $q$ & 5  & 5 \\
hidden state dimension $s_{h}$ & 256 & 256\\
capsule dimensions [$d_{u}, d_{r}$] & [16,16] & [16,16]\\
iterations $z$ & 3 & 3\\
sliding-margin $\gamma$ & 0.4 & 0.4\\
down-weighting $\lambda$ & 1.0 & 0.5\\
learning rate  & 0.001 & 0.001\\
dropout probability   & 0.0 & 0.7\\
L2 regularization strength  & 0.0001 & 0.0\\
\bottomrule
\end{tabular}
\caption{Parameter settings}
\label{table:parameters}
\end{table}

\subsection{Overall Performance}
\begin{figure}[htbp]
\centering
\includegraphics[width=3in]{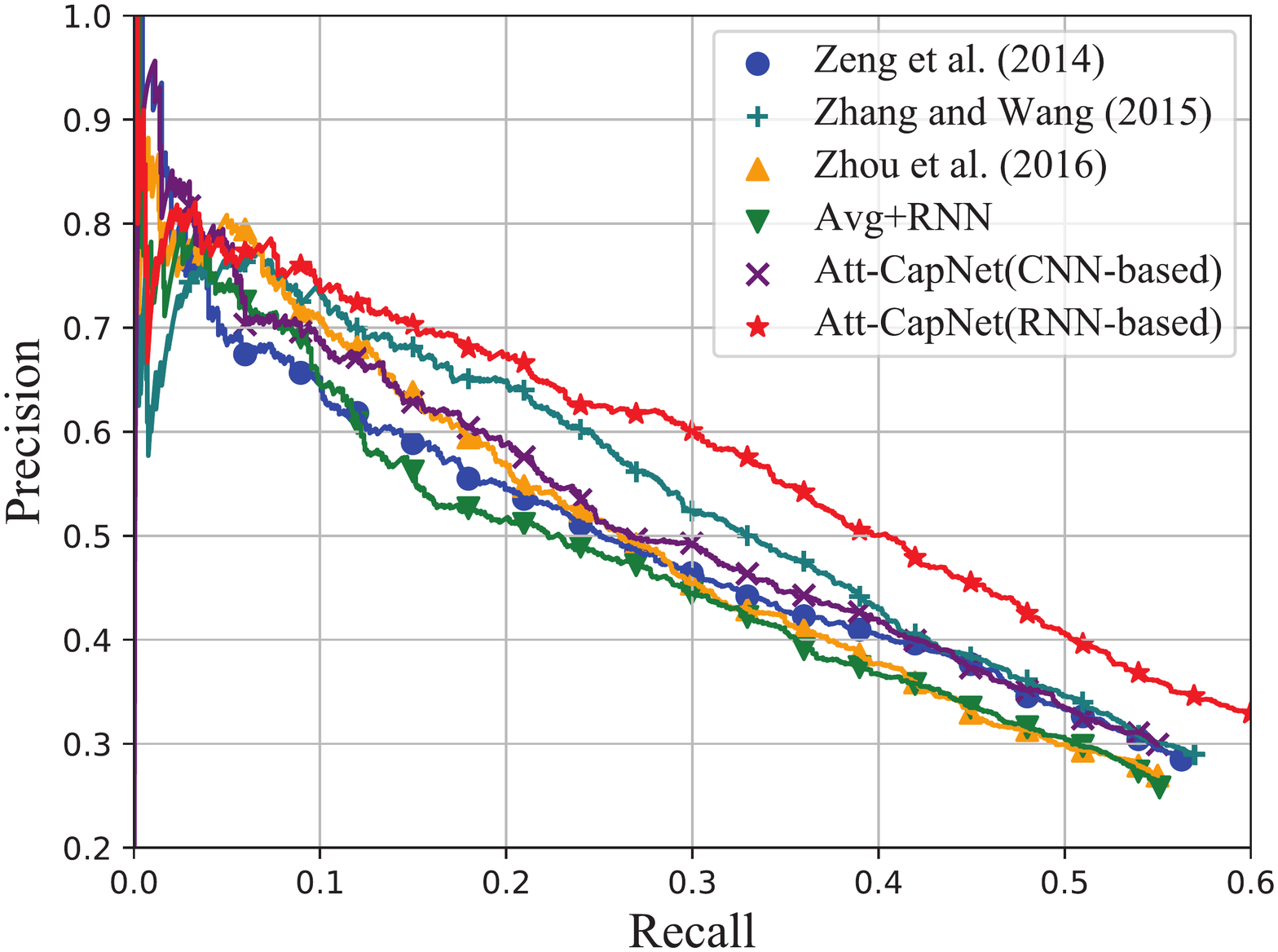}
\caption{The PR curves of all the baselines on NYT-10}
\label{fig:curve}
\end{figure}

We compare our method with the previous baselines on the two datasets respectively. For the dataset NYT-10, the performance of all the methods is compared in Figure~\ref{fig:curve}. The figure draws the PR curves of all the baselines. Apparently, we can see, 1) our Att-CapNet (RNN-based) model achieves the best PR curve, which outperforms the other baselines at nearly all range of the recall. 2) our Att-CapNet (CNN-based) model is slightly better than the method Max-pooling+CNN~\cite{zeng2014relation}. 3) RNN models tendentiously outperform CNN ones for their strong ability of extracting low-level relation features from the sequence. 4) our attentive capsule network is effective for relation extraction integrated with either CNN or RNN.

A detailed comparison of baselines with precision, recall, F1 and PR curve areas is shown in Table~\ref{table:op@nyt}, which indicates, 1) our Att-CapNet (RNN-based) obtains better results on all the indicators than the baselines and increases F1 scores by at least 3.2\% over the other baselines. 2) Att-CapNet (CNN-based) is slightly better than Max-pooling+CNN~\cite{zeng2014relation}. 3) the previous feature extractors such as RNN and CNN are all improved by integrating with our attentive capsule network.

\begin{table}[htbp]
\centering
\resizebox{\linewidth}{!}{
\begin{tabular}{rccccc}
\toprule
Methods & Precision(\%) & Recall(\%)  & F1(\%)  & PR\\
\toprule
\citeauthor{zeng2014relation}~\shortcite{zeng2014relation} & 28.5 & 56.3 & 37.8 & 0.35 \\
\citeauthor{zhang2015relation}~\shortcite{zhang2015relation}  & 28.9 & 57.0 & 38.4 & 0.34 \\
\citeauthor{zhou2016attention}~\shortcite{zhou2016attention}  & 26.9 & 54.9 & 36.1 & 0.34 \\
Avg+RNN  & 25.7 & 55.1 & 35.1 & 0.33 \\
\midrule
Att-CapNet (CNN-based) &  29.9 &  55.0 & 38.8 &  0.36 \\
Att-CapNet (RNN-based)   & \textbf {30.8} & \textbf {63.7} & \textbf {41.6} & \textbf{0.42} \\
\bottomrule
\end{tabular}}
\caption{Performance of all the baselines on NYT-10. PR represents precision-recall curve area.}
\label{table:op@nyt}
\end{table}

We further conduct paired t-test (10-fold, F1 score) to evaluate the statistical significance of our results in terms of p-value and confidence intervals. Table~\ref{table:significance@nyt} shows that all the p-values are less than 5.0e-02 and the increases in F1 score are at least 2.2\%. Therefore, all our performance improvements are statistically significant.

\begin{table}[htbp]
\centering
\resizebox{\linewidth}{!}{
\begin{tabular}{rcc}
\toprule
Baselines & p-value  & CI (Confident level 95\%) \\
\midrule
\citeauthor{zeng2014relation}~\shortcite{zeng2014relation} & 1.0e-02 & [0.023, 0.051] \\
\citeauthor{zhang2015relation}~\shortcite{zhang2015relation}  & 1.2e-02 & [0.022, 0.041] \\
\citeauthor{zhou2016attention}~\shortcite{zhou2016attention}  & 2.4e-05 & [0.052, 0.077] \\
Avg+RNN  & 1.1e-04 & [0.044, 0.064] \\
\bottomrule
\end{tabular}}
\caption{The statistical significance in the difference between Att-CapNet (RNN-based) and the baselines. CI represents confidence intervals.}
\label{table:significance@nyt}
\end{table}

From the results in Table~\ref{table:sem} on SemEval-2010 dataset\footnote{We compare our method with previous works which do not depend on the parser information and external data such as WordNet.}, we can have the following observations, 1) our Att-CapNet (RNN-based) outperforms all the other feature clustering methods. 2) Att-CapNet (CNN-based) is better than the traditional CNN model with max-pooling. 3) RNN models are better than CNN ones under the same settings. 4) our attentive capsule network is more useful than the other feature clustering methods such as max-pooling or word-level attention.

\begin{table}[htbp]
\centering
\resizebox{\linewidth}{!}{
\begin{tabular}{rccccc}
\toprule
Methods & Features & F1(\%)  \\
\toprule
\citeauthor{zeng2014relation}~\shortcite{zeng2014relation} & WE (dim=50) & 69.7 \\
\citeauthor{zeng2014relation}~\shortcite{zeng2014relation}$^\text{\ddag}$ & WE (dim=50)+PE & 79.8 \\
\citeauthor{zhang2015relation}~\shortcite{zhang2015relation}  & WE (dim=50) & 80.0 \\
\citeauthor{zhang2015relation}~\shortcite{zhang2015relation}  & WE (dim=300) & 82.5 \\
\citeauthor{zhang2015relation}~\shortcite{zhang2015relation}$^\text{\ddag}$  & WE (dim=50)+PE & 81.0 \\
\citeauthor{zhou2016attention}~\shortcite{zhou2016attention}  & WE (dim=50) & 82.5\\
\citeauthor{zhou2016attention}~\shortcite{zhou2016attention}  & WE (dim=100) & 84.0\\
\citeauthor{zhou2016attention}~\shortcite{zhou2016attention}$^\text{\ddag}$  & WE (dim=50)+PE & 81.7\\
Avg+RNN$^\text{\ddag}$  & WE (dim=50)+PE & 78.4 \\
\midrule
Att-CapNet (CNN-based) &  WE (dim=50)+PE &  80.4  \\
Att-CapNet (RNN-based) &  WE (dim=50)+PE & \textbf {84.5}   \\
\bottomrule
\end{tabular}}
\caption{Performance of all the baselines on SemEval-2010 Task 8. WE, PE respectively stand for word embedding and position embedding. Methods with $\ddag$ are our implementations. The other results are reported in their papers.}
\label{table:sem}
\end{table}

\subsection{Effect of Our Method on Multi-labeled Sentences}
To evaluate the effect of our method on the multi-labeled sentences, we randomly select 500 sentences, which have more than one labels, from NYT-10 for testing. The previous methods cannot predict multiple relations, and all of them can only obtain a low recall rate of about 0.40. Therefore, we define a threshold confident score for all the previous methods to make them predict multiple relations. We tune the threshold to ensure that all the previous methods can achieve maximum F1 scores\footnote{We compute F1 scores for a series of confidence scores and select the maximum ones for the previous methods. The interval of confidence scores is 0.1.}. As shown in Table~\ref{table:500sen}, our Att-CapNet (RNN-based) achieves the best precision, recall and F1. Our methods can recall more relation labels in the scenario where a sentence contains different relations.

\begin{table}[htbp]
\centering
\resizebox{\linewidth}{!}{
\begin{tabular}{rccccc}
\toprule
Methods & Precision(\%)  & Recall(\%)  & F1(\%)  \\
\toprule
Max-pooling+CNN & 88.4 & 91.9 & 90.1  \\
Max-pooling+RNN & 89.3 & 91.8 & 90.5 \\
Att+RNN  & 88.8 & 90.6 & 89.7 \\
Avg+RNN  & 86.9 & 90.5 & 88.6 \\
\midrule
Att-CapNet (CNN-based) &  87.3 &  93.0 & 90.1  \\
Att-CapNet (RNN-based)   & \textbf {89.9} & \textbf {93.7} & \textbf {91.8}  \\
\bottomrule
\end{tabular}}
\caption{Performance of all the baselines on selected 500 multi-labeled sentences from NYT-10.}
\label{table:500sen}
\end{table}

\begin{table*}[htbp]
\centering
\small
\begin{tabular}{|m{10cm}|m{0.8cm}<{\centering}|m{1.8cm}<{\centering}|m{0.5cm}<{\centering}|m{0.5cm}<{\centering}|m{1.5cm}<{\centering}|m{0pt}}
\cline{1-6}
\multirow{2}{*}[-5pt]{Sentences} & \multirow{2}{*}[-5pt]{Labels} & \multicolumn{4}{c|}{RNN} & \\[6pt]
\cline{3-6}
 &  & Max-pooling  & Avg. & Att. & Att-CapNet &\\[6pt]
\cline{1-6}
\multirow{2}{10cm}{\noindent{\bf S1}: Twenty years ago, another {\bf [Augusta]} native, {\bf [Larry Mize]}, shocked Greg Norman in a playoff by holing a 140-foot chip for birdie on the 11th hole to win the masters in a playoff.} & PB & 0 & 0 & 0 & 1 &\\ [6pt]
\cline{2-6}
 & PL & 0 & 1 & 1 & 1 &\\ [6pt]
\cline{1-6}
\multirow{2}{10cm}{\noindent{\bf S2}: Brothers or cousins except for its drummer, Oscar Lara, the band originally comes from {\bf [Sinaloa]}, {\bf [Mexico]}, but has lived in San Jose, California, for nearly 40 years.}& LC & 1 & 0 & 0 & 1 &\\[6pt]
\cline{2-6}
 & CA & 0 & 1 & 1 & 1 & \\[6pt]
\cline{1-6}
\multirow{3}{10cm}{\noindent{\bf S3}: The white house in April sharply criticized the speaker of the house, Nancy Pelosi, for visiting {\bf [Syria]}'s capital, {\bf [Damascus]}, and meeting with president Bashar Al-Assad, even going so far as calling the trip ``bad behavior'', in the words of vice president Dick Cheney.} & LC & 0 & 0 & 0 & 1 &\\[6pt]
\cline{2-6}
 & CA & 0 & 0 & 1 & 1 &\\[6pt]
\cline{2-6}
  & CC & 1 & 1 & 0 & 1 &\\[6pt]
\cline{1-6}
\end{tabular}
\caption{A case study of selective multi-labeled sentences for the four feature clustering methods based on RNN. The entities are labeled in the bold brackets. ``PB'', ``PL'', ``LC'', ``CA'' and ``CC'' are relation labels in the dataset, which are ``person/place\_birth'', ``person/place\_lived'',``location/contain'', ``country/administrative\_divisions'' and ``country/capital'' respectively.}
\label{table:cases}
\end{table*}

\subsection{Effect of Various Modules}
\begin{figure}[htbp]
\centering
\includegraphics[width=3in]{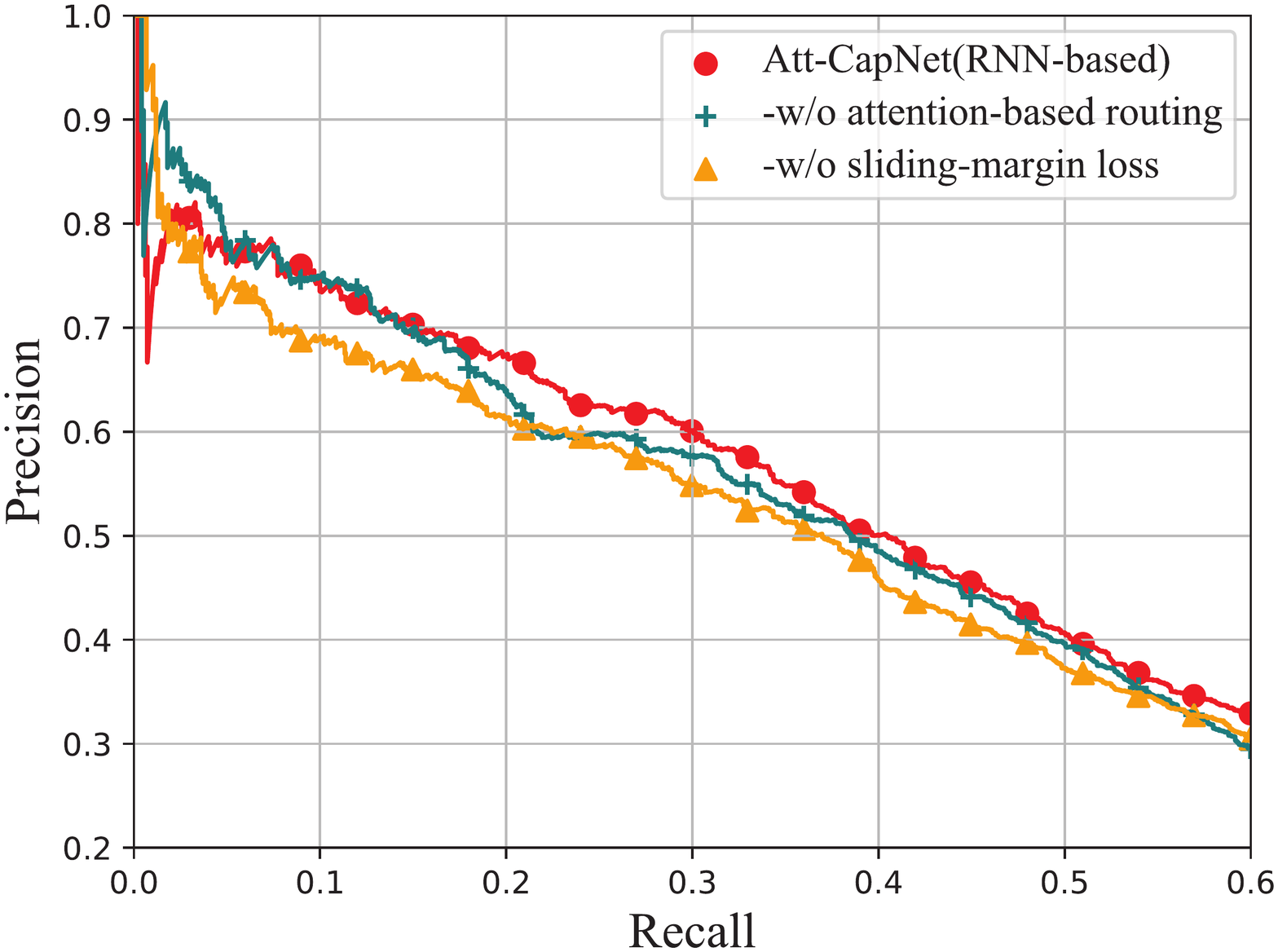}
\caption{The PR curves of the Att-CapNet (RNN-based) with various modules on NYT-10}
\label{fig:curve_various}
\end{figure}

In this subsection, we evaluate various modules of our method including attention-based routing algorithm and sliding-margin loss function. Our main method outperforms the other variants, although the variants may still prove useful when applied to other tasks. We apply our model Att-CapNet (RNN-based) and its two sub-models, which are without the attention-based routing algorithm (dynamic routing) and sliding-margin loss (fixed-margin loss), to the two datasets respectively. The results shown in Table~\ref{table:f1_various} and Figure~\ref{fig:curve_various} indicate, 1) our main model and the two variants are better than the best baseline feature clustering method under the same settings. 2) our attention-based routing algorithm and sliding-margin loss are both useful for capsule network, which significantly enhance the performance of relation extraction.
 
\begin{table}[htbp]
\centering
\resizebox{\linewidth}{!}{
\begin{tabular}{lccccc}
\toprule
Methods & Features & F1(\%)  \\
\toprule
\citeauthor{zhou2016attention}~\shortcite{zhou2016attention}  & WE (dim=50)+PE & 81.7\\
Att-CapNet (RNN-based) &  WE (dim=50)+PE & \textbf {84.5}   \\
 {-}w/o attention-based routing &  WE (dim=50)+PE & 83.6  \\
 {-}w/o sliding-margin loss &  WE (dim=50)+PE & 82.3   \\
\bottomrule
\end{tabular}}
\caption{Performance of Att-CapNet (RNN-based) with various modules on SemEval-2010 Task 8.}
\label{table:f1_various}
\end{table}

\subsection{Case Study}
We present practical cases in NYT-10 test set to show the effectiveness of our feature clustering method (Att-CapNet) compared to max-pooling, linear average (Avg.) and word-level attention (Att.) with the same feature extracting network (RNN). Table~\ref{table:cases} shows three multi-labeled sentences for relation extraction by all the four methods from which we can conclude that, 1) Att-CapNet method has recognized all the labeled relations. 2) the other three methods can only give one confident prediction. 3) a few methods even cannot recognize any relations such as max-pooling for the sentence \emph{S1}. 4) our feature clustering method is more capable to recognize highly overlapping relations.

\section{Conclusions and Future Work}
In this paper, we propose a novel capsule-based approach for multi-labeled relation extraction to handle the highly overlapping relations and improve the capability of clustering relation features. To our best knowledge, this is the first attempt that applies capsule network to solve the challenging task. The proposed model consists of a concise pipeline. First, we extract low-level semantic information with Bi-LSTM. Then, the low-level features are clustered to be high-level relation representations with attentive capsule network. Finally, sliding-margin loss is proposed to train the model reasonably with all the relations including the ``no relation''. Our experiments show that the proposed approach achieves significant improvement for multi-labeled relation extraction over previous state-of-the-art baselines.

In future, our solutions of features clustering can be generalized to other tasks that deal with overlapping and discrete features. For instance, a possible attempt might be to perform reading comprehension.

\section*{Acknowledgments}
This work is supported by National China 973 Project No. 2015CB352401; Chinese National Research Fund (NSFC) Key Project No. 61532013 and No. 61872239. FDCT/0007/2018/A1, DCT-MoST Joint-project No. (025/2015/AMJ), University of Macau Grant Nos: MYRG2018-00237-RTO, CPG2018-00032-FST and SRG2018-00111-FST of SAR Macau, China. National Key Research and Development Program of China (No. 2017YFB0304100), National Natural Science Foundation of China (No. 61672343 and No. 61733011), Key Project of National Society Science Foundation of China (No. 15-ZDA041), The Art and Science Interdisciplinary Funds of Shanghai Jiao Tong University (No. 14JCRZ04).

\bibliography{capsule}
\bibliographystyle{aaai19}

\end{document}